\documentclass[runningheads]{llncs}
\usepackage{graphicx}
\usepackage{hyperref}
\usepackage{amsmath}
\begin{document}
\title{Improving the Reliability of Semantic Segmentation of Medical Images by Uncertainty Modeling with Bayesian Deep Networks and Curriculum Learning%\thanks{Supported by Hiroshima University.}
}
\titlerunning{Improving the Reliability of Semantic Segmentation 
%by Uncertainty Modelling
}
% If the paper title is too long for the running head, you can set
% an abbreviated paper title here
%
\author{Sora Iwamoto\inst{1}%\orcidID{0000-0001-9523-2047} 
\and Bisser  Raytchev\inst{1}\thanks{corresponding author, email: bisser@hiroshima-u.ac.jp}
\and Toru Tamaki\inst{2}
\and Kazufumi Kaneda\inst{1} %inst{2,3}\orcidID{0000-0002-2146-415X}
}
\authorrunning{Iwamoto, Raytchev, Tamaki and Kaneda}
% First names are abbreviated in the running head.
% If there are more than two authors, 'et al.' is used.
%
\institute{Graduate School of Advanced Science and Engineering, Hiroshima University%Affiliations\\%Hiroshima University, Hiroshima Prefecture 739-8525, Japan 
%\email{email}%lncs@hiroshima-u.ac.jp}\\
%\url{https://www.hiroshima-u.ac.jp/}
\and Department of Computer Science, Nagoya Institute of Technology
}
\maketitle              % typeset the header of the contribution
\begin{abstract}
In this paper we propose a novel method which leverages the uncertainty measures provided by Bayesian deep networks through curriculum learning so that the uncertainty estimates are fed back to the system to resample the training data more densely in areas where uncertainty is high. We show in the concrete setting of a semantic segmentation task (iPS cell colony segmentation) that the proposed system is able to increase significantly the reliability of the model.

\end{abstract}
\section{Introduction}
Although in recent years deep neural networks have achieved state-of-the-art performance on many medical image analysis tasks, even surpassing human-level performance in certain cases \cite{skin_cancer_nature}, \cite{retinopathy_JAMA2016}, their extensive adoption in clinical settings has been hampered by their false over-confidence when confronted with out-of-distribution (OOD) test samples (samples that lie far away from the data which they have been trained with). This is due to the fact that the probability vector obtained from the softmax output is erroneously interpreted as model confidence \cite{gal2015dropout}. It may be amusing if a deep net trained with cats and dogs images classifies a human as a dog with $98\%$ probability, but similar mistake due to encountering test samples lying outside the data distribution of a cancer detection system can lead to  life-threatening situations, thus the reluctance of some medical professionals to adopt such systems wholeheartedly. 

In order to address this problem different risk-aware Bayesian networks have been proposed \cite{BNN}, \cite{gal2015dropout}, \cite{mobiny2019risk}, which rather than point estimates, as in popular deep learning models, are able to output uncertainty estimates, which can provide information about the \emph{reliability} of the trained models, thus allowing the users to take necessary actions to ensure safety when the model is under-confident or falsely over-confident. 

In this paper, we endeavor to take this work one step further, i.e. not merely to provide uncertainty measures, but to leverage those through curriculum learning \cite{curriculum}, so that the uncertainty estimates are fed back to the system to resample the training data more densely in areas where the uncertainty is high. We show, in the setting of a concrete semantic segmentation task, that the reliability of the model can be significantly increased without decreasing segmentation accuracy, which can potentially lead to wider acceptance of deep learning models in clinical settings where safety is first priority.    

\begin{figure}[t]
\includegraphics[width=\textwidth]{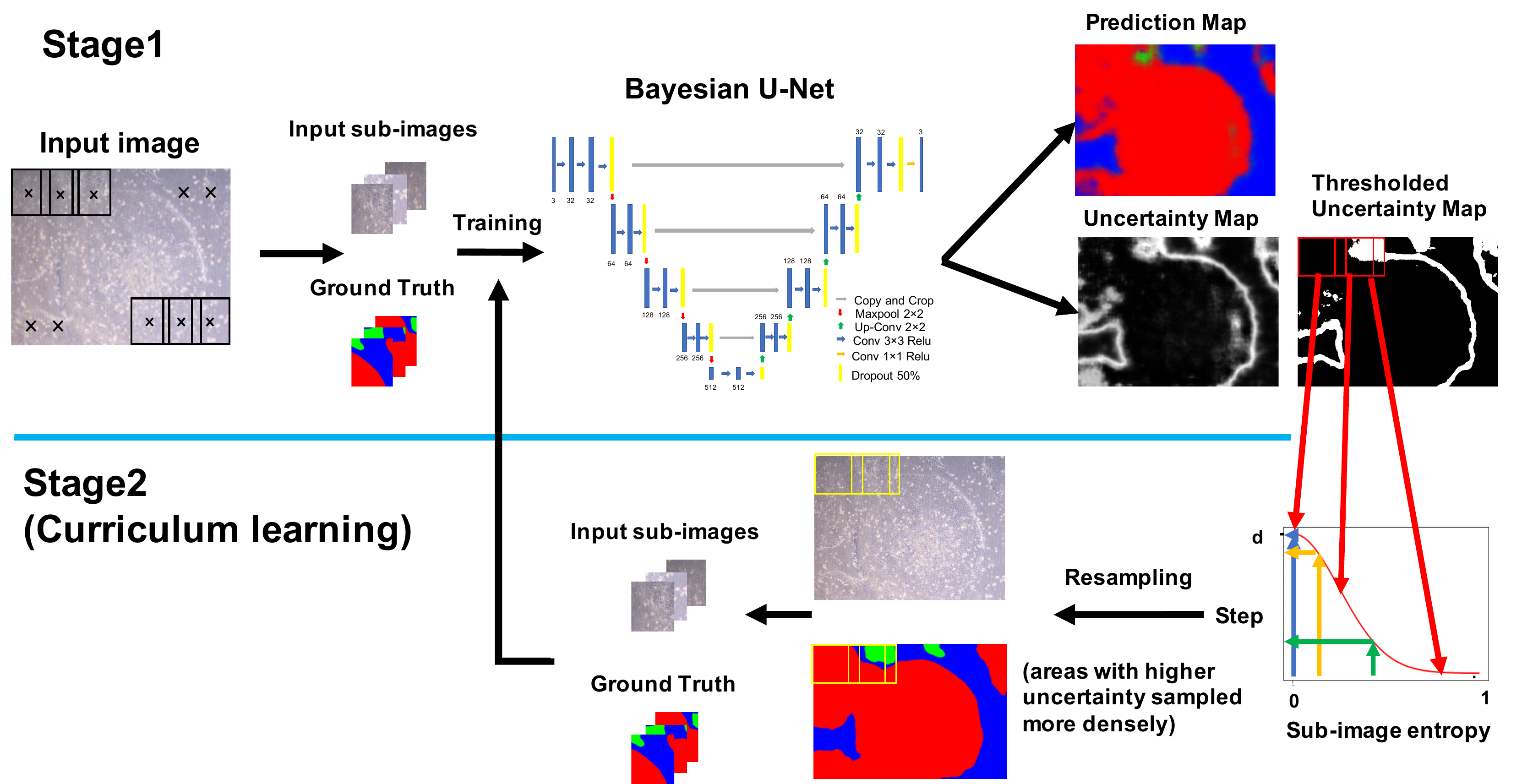}
\caption{Overview of the proposed system.} \label{fig1}
\end{figure}

\section{Methods}
Fig.~\ref{fig1} provides an overview of the proposed method which consists of two stages.
During stage 1 a large image which needs to be segmented (the class of each pixel needs to be determined) is input into the system. We assume large bio-medical images with predominantly local texture information (one example would be colonies of cells), where local texture statistics are more important for the correct segmentation and distant areas might not be correlated at all.
We extract sub-images of size $d \times d$ pixels from the large image, and these are sent to a Bayesian U-Net for learning the segmentation end-to-end, using ground truth segmentation map images provided by experts. In the lack of additional information the sub-images can be extracted from the large image in a sliding window manner, scanning the large image left-to-right and top-to-bottom using a predefined step $s$.% (see Fig.~\ref{fig1}). %as shown in Fig.~\ref{fig1}. 

To obtain the model's uncertainty of its prediction, we apply Monte Carlo (MC) Dropout \cite{gal2015dropout}, \cite{devries2018leveraging}, \cite{mobiny2019risk} to a U-Net \cite{ronneberger2015unet}.  
An approximate predictive posterior distribution can be obtained by Monte Carlo sampling over the network parameters by keeping the dropout mechanism \cite{drop} at test time and performing the prediction multiple times during the forward pass (dropout approximately integrates over the model's weights \cite{gal2015dropout}). The predictive mean %$\mu_{pred}$ 
for a test sample $\mathbf{x}^{*}$ is  
\begin{equation}
\mu_{p r e d} \approx \frac{1}{T} \sum_{t=1}^{T} p\left(y^{*} \mid \mathbf{x}^{*}, \hat{\mathbf{w}}_{t}\right)
\end{equation}
where $T$ is the number of MC sampling iterations, $\hat{\mathbf{w}}_t$ represents the network weights with dropout applied to the units at the $t$-th MC iteration, and $y^{*}$ is class vector. For each test sample  $\mathbf{x}^{*}$, which is a pixel in an input sub-image, the prediction is chosen to be the class with largest predictive mean (averaged for overlapping areas across sub-images). In this way a Prediction Map for the whole image is calculated, as shown in Fig.~\ref{fig1}.  

To quantify the \emph{model uncertainty} we adopt predictive entropy $H$ as proposed in \cite{entropy}:
\begin{equation}
H\left(\mathbf{y}^{*} \mid \mathbf{x}^{*}, \mathcal{D}\right)=-\sum_{c} p\left(y^{*}=c \mid \mathbf{x}^{*}, \mathcal{D}\right) \log p\left(y^{*}=c \mid \mathbf{x}^{*}, \mathcal{D}\right)
\end{equation}
where $c$ ranges over the classes. Since the range of the uncertainty values can vary across different datasets $\mathcal{D}$ or models, similarly to \cite{mobiny2019risk} we adopt the normalized entropy $H_{\text {norm}} \in [0, 1]$, computed as 
$H_{\text {norm}}=\frac{H-H_{\min }}{H_{\max }-H_{\min }}$. In this way, an Uncertainty Map is calculated for the whole image, which can be thresholded using a threshold $H_T$ to obtain the \emph{Thresholded Uncertainty Map} (see Fig.~\ref{fig1}) where each pixel's prediction is considered \emph{certain} if the corresponding value in the map is larger than $H_T$ and \emph{uncertain} otherwise.  

In stage 2 of the proposed method, we utilize Curriculum Learning \cite{curriculum} to leverage the information in the Uncertainty Map about the uncertainty of the model to improve its reliability (methods for evaluating the reliability of a model are explained in Section 3). The main idea is that \textbf{\emph{areas where segmentation results are uncertain need to be sampled more densely than areas where the model is certain}}. The uncertainty $H(S)$ of sub-images $\cal{S}$ is calculated as
%  \begin{equation}
%    H(S) = - \frac{1}{N}\sum_{i \in S}\sum_{c} p^{(i)}_{c} \log{p^{(i)}_{c}},
%    \label{equ:entropy}
%  \end{equation}
%which represents 
the average entropy obtained from the uncertainty values in the Uncertainty Map 
\emph{pmf}s $p^{(i)}$ for each pixel (indexed by $i$) 
corresponding to the sub-image. %, and $c$ indexes the classes. 
Using $H(S)$ as a measure of the current sub-image's uncertainty, the position
of the next location where to resample a new sub-image is given by
  \begin{equation}
    f(H(S)) = d \exp\{{-(H(S))^2/2\sigma^2}\}.
    \label{equ:gauss}
  \end{equation}
where $d$ is the size of the sub-image and $\sigma$ the width of the Gaussian. This process is illustrated in Fig.~\ref{fig1}, starting at the upper left corner of the input image, uncertainty for the current sub-image is calculated
%using Eq.~\ref{equ:entropy}, 
and the step size in pixels to move in the horizontal direction
is calculated by Eq.~\ref{equ:gauss}. The whole image is resampled this way to re-train the model and this process (stage 2) can be repeated several times until no further improvement in reliability is obtained.

Additionally, we propose a second method (Method 2, single-staged), which does not use curriculum learning, i.e. consists of a single training stage. This method, rather than resampling the training set, directly uses the values in the Uncertainty Map to improve model reliability. This method initially trains the Bayesian U-Net for 5 epochs using cross-entropy loss (several other losses have also been tried as shown in the experiments), after which generates an Uncertainty Map similarly to the curriculum learning based method, and continues the training for 5 more epochs augmenting the training loss with a term which tries directly to minimize the uncertainty values of the Uncertainty Map - for this reason we call it \textbf{\emph{Uncertainty Loss}}.
Note that the term that minimizes the uncertainty is added to the cross-entropy term \emph{after} the cross-entropy loss has been minimized for several epochs, so that this does not encourage overconfident false predictions.

%\begin{figure}
%\includegraphics[width=\textwidth]{fig1.eps}
%\caption{A figure caption is always placed below the illustration.
%Please note that short captions are centered, while long ones are
%justified by the macro package automatically.} \label{fig1}
%\end{figure}

\section{Experiments}

\begin{figure}
\includegraphics[width=\textwidth]{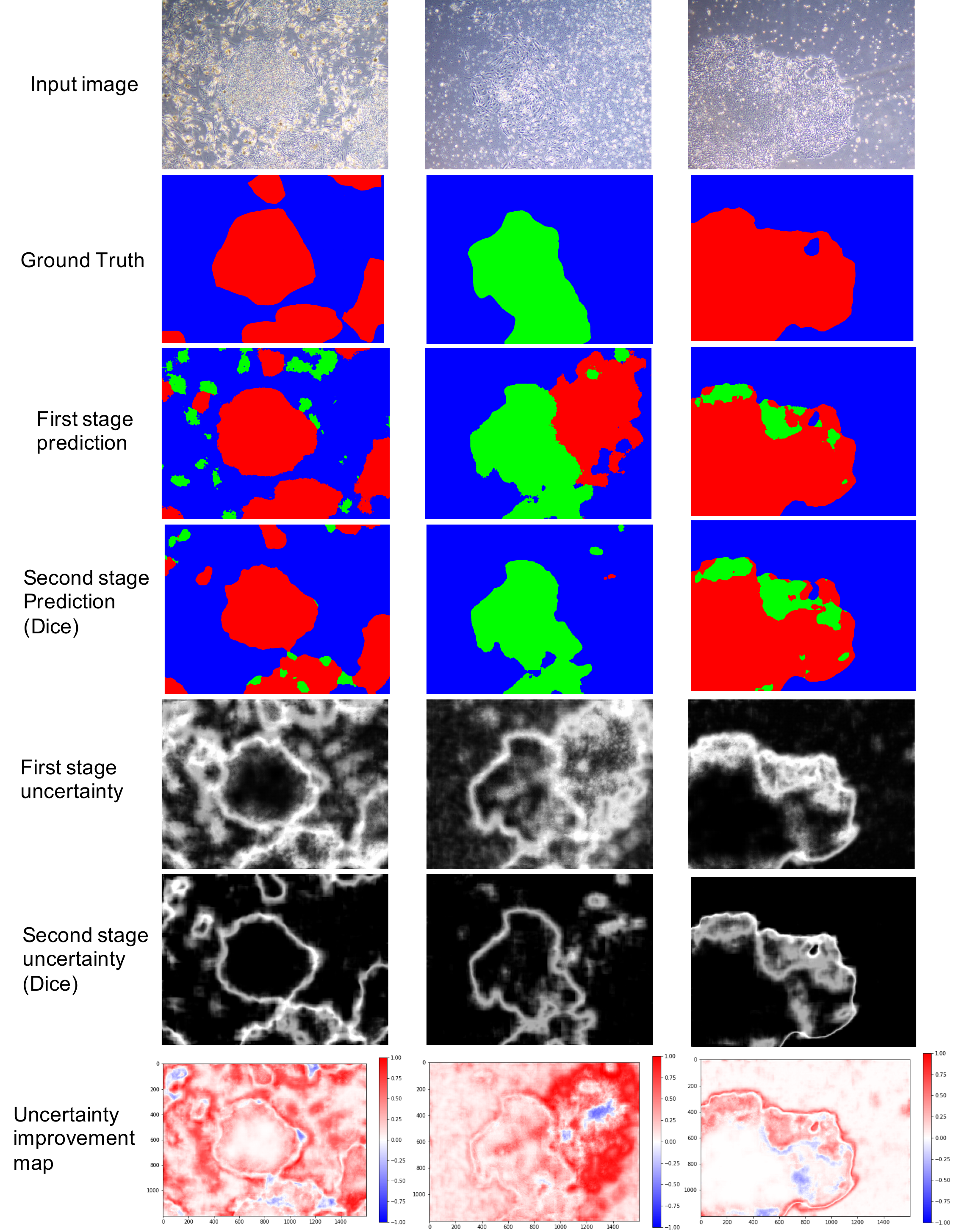}
\caption{Segmentation results and uncertainty maps for several images from the iPS dataset. In the segmentation results (2nd to 4th rows from the top) red corresponds to Good colonies, green to Bad colonies and blue to the culture medium. In the uncertainty maps (5th and 6th rows)  high uncertainty level is represented by high intensity values. First two columns show instances where both segmentation accuracy and reliability was improved significantly by curriculum learning (4th and 6th rows) in comparison with the baseline method (3rd and 5th rows), while the last column shows a case without much improvement. The last row is a heat map where reduction in uncertainty between first and second stage of learning is shown in red and increase in blue. (Best viewed in color)} \label{fig2}
\end{figure}

In this section we evaluate the performance of the proposed methods on a dataset
which consists of 59 images showing colonies of undifferentiated and
differentiated iPS cells obtained through phase-contrast microscopy. 
%Induced pluripotent stem (iPS) cells \cite{ips}, for whose discovery S. Yamanaka received the 
%Nobel prize in Physiology and Medicine in 2012, contain great promise for regenerative medicine.
%However, to make it practical, a
%steady supply of iPS cells obtained through harvesting of individual cell colonies
%is needed and automating the detection of abnormalities arising during the cultivation process is crucial.
The task we have to solve is to segment the input images into three categories: Good (undifferentiated), Bad (differentiated) and Background (BGD, the culture medium).
  Several representative images together with ground-truth provided by experts can be seen in 
  Fig.~\ref{fig2}.
  All images in this dataset are of size $1600 \times 1200$ pixels. 
  %Several images contained a few locations where even the experts were not sure what the corresponding class was.
  %Such ambiguous regions are shown in pink and these areas were not used during training and not evaluated during test. \\

{\bf Network Architecture and Hyperparameters:} 
We used a Bayesian version of U-Net \cite{mobiny2019dropconnect}, \cite{mobiny2019risk},  \cite{ronneberger2015unet}, the architecture of which can be seen in Fig.~\ref{fig1}, with $50\%$ dropout applied to all layers of both the encoder and decoder parts.

  The learning rate was set to $1e-4$ for the Stage 1 learning, and to $1e-6$ for the curriculum learning (beyond stage 2). For the optimization procedure we used ADAM \cite{adam} ($\beta_1=0.9, \beta_2=0.999$),
  batch size was 12, training for 10 epochs per learning stage, while keeping the model weights corresponding to minimal loss on the validation sets. All models were implemented using Python 3.6.10, TensorFlow 2.3.0 \cite{tensorflow2015-whitepaper} and Keras 2.3.1 \cite{keras}.  Computation was performed %on a computer with Intel(R) Xeon(R) CPU E5-2620 0 @ 2.00GHz, 
  using NVIDIA GeForce GTX1080 Ti.
  %Code will be made publicly available on github after paper acceptance.
  Code is available from \url{https://github.com/sora763/Uncertainty}. 
  %githubのリンクはhttps://github.com/sora763/である.(採択後公開)
The size of the sub-images was fixed to $d=160$ (i.e. $160 \times 160$ pixels). The width of the Gaussian in Eq.~\ref{equ:gauss} was empirically set to $\sigma = 0.4$ for all experimental results. The values of the other system parameters used in section 2 were set to $T = 10$, $s = 10$, $H_T = 0.5$ throughout the experiments. 

{\bf Evaluation procedure and criteria:} Evaluation was done through 5-fold cross validation by splitting the data into training, validation and test sets in proportions 3:3:1. Each method was evaluated by using the metrics described below, which are adopted from \cite{mobiny2019dropconnect}. 
In a Bayesian setting there are four different possible cases for an inference: it can be (a) incorrect and uncertain (True Positive, TP); (b) correct and uncertain (False Positive, FP); (c) correct and certain (True Negative, TN); (d) incorrect and certain (False Negative, FN). Correctness can be obtained by comparing the Prediction Map with the Ground Truth, while certain/uncertain values can be obtained from the Thresholded Uncertainty Map described in section 2.  

\begin{enumerate}
\item Negative Predictive Value (NPV)\\
The model should predict correctly if it is certain about its prediction. This can be evaluated by the following conditional probability and corresponds to the NPV measure in a binary test:
\begin{eqnarray}
P(\text { correct } \mid \text { certain })&=\frac{P(\text { correct, certain })}{P(\text { certain })}%\\
&=\frac{\mathrm { TN }}{\mathrm { TN+FN }}
\end{eqnarray}
\item True Positive Rate (TPR) \\
The model should be uncertain if the prediction is incorrect. This can be evaluated by the following conditional probability and corresponds to the TPR measure in a binary test:

\begin{eqnarray}
P(\text { uncertain } \mid \text { incorrect })&=\frac{P(\text { uncertain, incorrect })}{P(\text { incorrect })}%\\
&=\frac{\mathrm{TP}}{\mathrm{TP}+\mathrm{FN}}
\end{eqnarray}
\item Uncertainty Accuracy (UA)\\
Finally, the overall accuracy of the uncertainty estimation can be measured as the ratio of the desired cases (TP and TN) over all possible cases:

\begin{eqnarray}
  \mathrm{UA}=\frac{\mathrm{TP}+\mathrm{TN}}{\mathrm{TP}+\mathrm{TN}+\mathrm{FP}+\mathrm{FN}}
\end{eqnarray}
\end{enumerate}
For all metrics described above, higher values indicate a model that performs better. Additionally, overall segmentation performance was evaluated using the mean Intersection-over-Union (IoU, or Jaccard index), as is common for segmentation tasks. For each score the average and standard deviation obtained from 5-fold cross-validation are reported. \\

{\bf Experimental results for the curriculum learning:} Table~\ref{ips_result} shows the results obtained by the proposed method based
on resampling with curriculum learning, compared with the single-stage baseline (first row) using the same U-Net trained with cross-entropy (CE) loss without uncertainty modelling through curriculum learning. 
The second row in the table reports results obtained with curriculum learning using cross-entropy loss over 4 learning stages (first same as the baseline and next 3 stages using curriculum learning). Third row corresponds to results using Dice loss \cite{dice} for the second stage (curriculum learning), fourth row corresponds to results using the Sensitivity-Specificity (SS) loss \cite{SSLoss} for the second stage (curriculum learning), and the last row corresponds to results using both cross-entropy and Dice loss for the second stage (curriculum learning).
Regarding the system reliability evaluation, the results show that a big improvement is achieved when using the proposed curriculum learning method: in the case when Dice loss is used which achieves best performance, TPR improved by $12\%$, NPV by $9\%$ and UA by $6\%$. It was found that when using Dice loss, SS loss or CE+Dice loss for the curriculum learning necessitates only a single stage of curriculum learning (no significant improvement observed after that), while if the original CE loss is used 3 stages of curriculum learning were needed for best results (note that using Dice loss instead of CE during the first stage resulted in much inferior results). Additionally, segmentation performance (as measured by IoU) also improved by about $1\%$ on average for all curriculum learning method compared with the baseline.

Fig.~\ref{fig2} shows segmentation results and uncertainty maps for several images from the iPS dataset. The first two columns show instances where both segmentation accuracy and reliability was improved significantly by curriculum learning in comparison with the baseline method, while the last column shows a case without much improvement. 

\begin{figure}
\includegraphics[width=\textwidth]{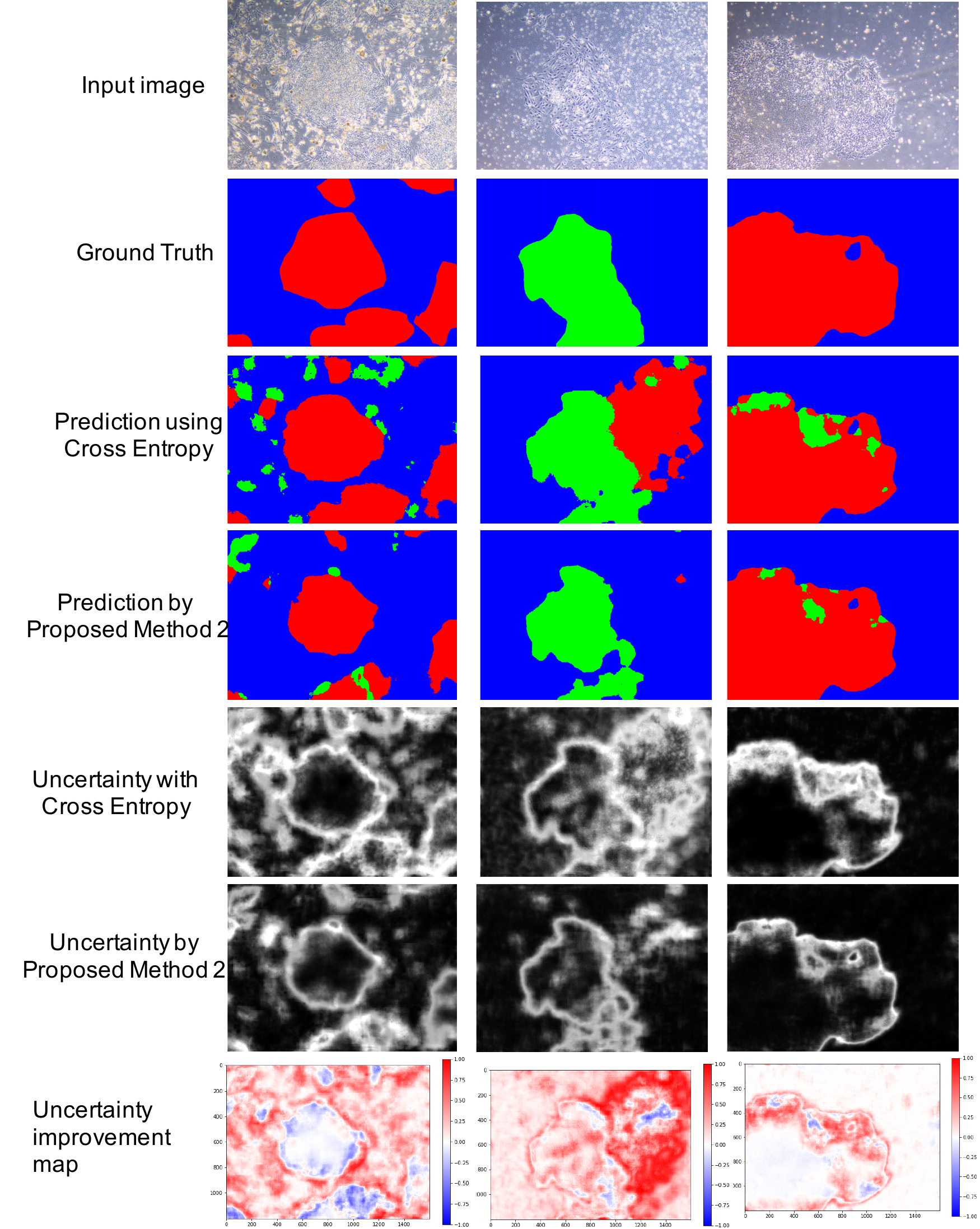}
\caption{Comparison of segmentation results and uncertainty maps for a baseline method using cross-entropy without uncertainty modelling (3rd row and 5th row) and proposed single-stage Method 2 (4th row and 6th row). The last row is a heat map where reduction in uncertainty between baseline method and Method 2 is shown in red and increase in blue. (Best viewed in color).} \label{fig3}
\end{figure}

%\begin{table}
%\caption{Table captions should be placed above the
%tables.}\label{tab1}
%\begin{tabular}{|l|l|l|}
%\hline
%Heading level &  Example & Font size and style\\
%\hline
%Title (centered) &  {\Large\bfseries Lecture Notes} & 14 point, bold\\
%1st-level heading &  {\large\bfseries 1 Introduction} & 12 point, bold\\
%2nd-level heading & {\bfseries 2.1 Printing Area} & 10 point, bold\\
%3rd-level heading & {\bfseries Run-in Heading in Bold.} Text follows & 10 point, bold\\
%4th-level heading & {\itshape Lowest Level Heading.} Text follows & 10 point, italic\\
%\hline
%\end{tabular}
%\end{table}

\begin{table}[t]
    \caption{Experimental results comparing reliability (NPV, TPR and UA) and segmentation accuracy (IoU) for baseline (first row) and curriculum learning based methods using different loss functions (second to fifth rows). }%iPS画像におけるCurriculum学習でのでの不確実性とSemantic Segmentationの推定結果. }
    \label{ips_result}
    \begin{center}
    \begin{tabular}{|c|c|c|c|c|c|}
        \hline
        Stage  & Loss &  NPV  & TPR  &  UA & IoU\\
        \hline
        \hline

        1 &  CE & 0.867 $\pm$ 0.026  & 0.313 $\pm$ 0.021   &  0.859 $\pm$ 0.027 & 0.783 $\pm$ 0.039  \\
        4 &  CE & 0.917 $\pm$ 0.021  & 0.352 $\pm$ 0.009   &  0.899 $\pm$ 0.022 & 0.796 $\pm$ 0.039\\
        2 &  Dice & {\bfseries0.955 $\pm$ 0.010}  & {\bfseries0.431 $\pm$ 0.011}   &  {\bfseries0.921 $\pm$ 0.015} & 0.794 $\pm$ 0.043\\
        2 &  SS & 0.894 $\pm$ 0.020  & 0.326 $\pm$ 0.015   &  0.883 $\pm$ 0.021 & {\bfseries0.797 $\pm$ 0.037}\\
        2 &  CE+Dice & 0.904 $\pm$ 0.019  & 0.340 $\pm$ 0.010   &  0.890 $\pm$ 0.020 & {\bfseries0.797 $\pm$ 0.042}   \\
        \cline{1-1} \cline{2-4}
        \hline
    \end{tabular}
    \end{center}
\end{table}

\begin{table}[!h]
    \caption{Experimental results comparing reliability (NPV, TPR and UA) and segmentation accuracy (IoU) for three baseline single-stage learning methods using different loss functions (first 3 rows) and the proposed single-stage method (Uncertainty Loss).}
    %等間隔にpatchを切り出したデータセットでのでの不確実性とSemantic Segmentationの推定結果. }
    \label{ips_result4}
    \begin{center}
    
    \begin{tabular}{|c|c|c|c|c|c|}
        \hline
        Stage  & Loss &  NPV  & TPR  &  UA & IoU\\
        \hline
        \hline

        1 &  CE & 0.867 $\pm$ 0.026  & 0.313 $\pm$ 0.021   &  0.859 $\pm$ 0.027 & 0.783 $\pm$ 0.039\\
        1 & SS &  0.913 $\pm$ 0.023 & 0.356 $\pm$ 0.013 & 0.896 $\pm$ 0.025 & 0.791 $\pm$ 0.039\\
        1 & CE+Dice & 0.913 $\pm$ 0.016 & 0.358 $\pm$ 0.021 & 0.895 $\pm$ 0.019 & 0.792 $\pm$ 0.042\\
        1 & Uncertainty Loss & {\bfseries0.935 $\pm$ 0.018} & {\bfseries0.382 $\pm$ 0.025} &  {\bfseries0.910 $\pm$ 0.018} & {\bfseries0.798 $\pm$ 0.041}\\

        \cline{1-1} \cline{2-4}
        \hline
    \end{tabular}
    \end{center}
\end{table}

{\bf Experimental results for the single-stage learning:} Table~\ref{ips_result4} shows the results obtained by the second proposed method (shown in the last row of the table), compared with three different single-stage baseline methods (first 3 rows) using the same U-Net trained with cross-entropy (CE) loss without uncertainty modelling (row 1 in the table), 
Sensitivity-Specificity (SS) loss without uncertainty modelling (row 2), and using both cross-entropy and Dice loss without uncertainty modelling (row 3 in the table). The results indicate that the proposed method outperforms all three baseline methods both in terms of reliability and segmentation accuracy. However, regarding reliability performance, this method did not perform as well as the curriculum learning based method. 

Fig.~\ref{fig3} shows segmentation results and uncertainty maps for the same images shown in Fig.~\ref{fig2}, this time comparing the proposed single-stage Method 2 with the baseline using cross-entropy without uncertainty modeling. Here again can be seen that modeling uncertainty leads to significant improvement in both segmentation accuracy and decrease in uncertainty compared with the baseline. 

%\begin{figure}
%\includegraphics[width=\textwidth]{images/pro2_m1.pdf}
%\caption{A figure caption is always placed below the illustration.
%Please note that short captions are centered, while long ones are
%justified by the macro package automatically.} \label{fig4}
%\end{figure}

\section{Conclusion}
Experimental results have shown that the proposed method was able to increase significantly the reliability of the segmentation model in the concrete setting of iPS cell colony segmentation. Further work includes application to alternative datasets and evaluation whether a hybrid model between both proposed method could lead to even further increase in reliability. 
%
%\cite{BNN}
%
% ---- Bibliography ----
%
% BibTeX users should specify bibliography style 'splncs04'.
% References will then be sorted and formatted in the correct style.
%
\bibliographystyle{splncs04}
\bibliography{reference}

\end{document}